\documentclass[lettersize,journal]{IEEEtran}

\IEEEoverridecommandlockouts  

\usepackage{amsmath,amsfonts}
\usepackage{algorithmic}
\usepackage{algorithm}
\usepackage{array}
\usepackage[caption=false,font=normalsize,labelfont=sf,textfont=sf]{subfig}
\usepackage{textcomp}
\usepackage{stfloats}
\usepackage{url}
\usepackage{verbatim}
\usepackage{graphicx}
\usepackage{cite}
\begin{document}

\title{Tool-MAD: A Multi-Agent Debate Framework for Fact Verification with Diverse Tool Augmentation and Adaptive Retrieval}

\author{
Seyeon Jeong\textsuperscript{1},
Yeonjun Choi\textsuperscript{1},
JongWook Kim\textsuperscript{2},
Beakcheol Jang\textsuperscript{1}\\
\textsuperscript{1}Graduate School of Information, Yonsei University, Seoul, Republic of Korea\\
\textsuperscript{2}Department of Computer Science, Sangmyung University, Seoul, Republic of Korea%
\thanks{This work was supported by the National Research Foundation of Korea (NRF) funded by Korean Government under Grant RS-2023-00273751.}%
}

\maketitle

\begin{abstract}
Large Language Models (LLMs) suffer from hallucinations and factual inaccuracies, especially in complex reasoning and fact verification tasks. Multi-Agent Debate (MAD) systems aim to improve answer accuracy by enabling multiple LLM agents to engage in dialogue, promoting diverse reasoning and mutual verification. However, existing MAD frameworks primarily rely on internal knowledge or static documents, making them vulnerable to hallucinations. While MADKE introduces external evidence to mitigate this, its one-time retrieval mechanism limits adaptability to new arguments or emerging information during the debate. To address these limitations, We propose Tool-MAD, a multi-agent debate framework that enhances factual verification by assigning each agent a distinct external tool, such as a search API or RAG module. Tool-MAD introduces three key innovations:
(1) a multi-agent debate framework where agents leverage heterogeneous external tools, encouraging diverse perspectives, (2) an adaptive query formulation mechanism that iteratively refines evidence retrieval based on the flow of the debate, and (3) the integration of Faithfulness and Answer Relevance scores into the final decision process, allowing the Judge agent to quantitatively assess the coherence and question alignment of each response and effectively detect hallucinations. Experimental results on four fact verification benchmarks demonstrate that Tool-MAD consistently outperforms state-of-the-art MAD frameworks, achieving up to 5.5\% accuracy improvement. Furthermore, in medically specialized domains, Tool-MAD exhibits strong robustness and adaptability across various tool configurations and domain conditions, confirming its potential for broader real-world fact-checking applications.
\end{abstract}

\begin{IEEEkeywords}
LLM, Multi-Agent Debate, Fact Verification.
\end{IEEEkeywords}

\section{Introduction}
\IEEEPARstart{L}{arge} Language Models (LLMs) have recently achieved strong performance across various NLP tasks~\cite{brown2020language,thoppilan2022lamda,driess2023palm}, such as dialogue generation, summarization, and knowledge extraction. However, they often suffer from hallucination—producing confident yet factually incorrect content~\cite{li2024dawn}. To mitigate this issue, prompt-based single-agent methods such as Chain-of-Thought (CoT) \cite{wei2022chain} and self-reflection \cite{shinn2023reflexion} guide reasoning or incorporate external tools like the Wikipedia API. While simple and effective, these approaches often fail to revise incorrect answers, a limitation referred to as \textit{Degeneration of Thought}~\cite{liang-etal-2024-encouraging}.

Recently, multi-agent debate frameworks have been proposed to overcome this by enabling multiple LLMs to engage in argumentation and mutual verification~\cite{du2023improving, liang-etal-2024-encouraging}. While promising, existing methods typically rely on static documents or internal model knowledge and still base final decisions on LLM-generated debate history, which remains vulnerable to hallucination. To enhance factual grounding, MADKE~\cite{wang2025learning} introduced static external retrieval before the debate. However, this evidence remains fixed during discussion, limiting the agents’ ability to adapt to new claims or knowledge gaps that emerge throughout the debate.

A key challenge underlying these limitations is that fact verification in practical scenarios rarely depends on a single, uniform source of evidence. Different external tools provide distinct advantages: search APIs offer broad coverage and access to real-time information, whereas Retrieval-Augmented Generation (RAG) modules retrieve semantically aligned and context-rich documents from curated corpora such as Wikipedia. Tasks involving breaking news or fast-evolving events require up-to-date external information, while tasks involving historically stable knowledge demand high-precision retrieval. Relying on a single type of retrieval often leads to systematic blind spots. Moreover, prior debate frameworks typically assume that evidence retrieved before the discussion is sufficient for the entire process, overlooking the dynamic nature of argumentation. In human debates, participants routinely refine their claims and gather new evidence when confronted with counterarguments or inconsistencies. Without a mechanism for iterative retrieval, debate agents remain restricted by the limitations of their initial evidence set.

\begin{figure}[t]
\centering
\includegraphics[width=0.9\columnwidth]{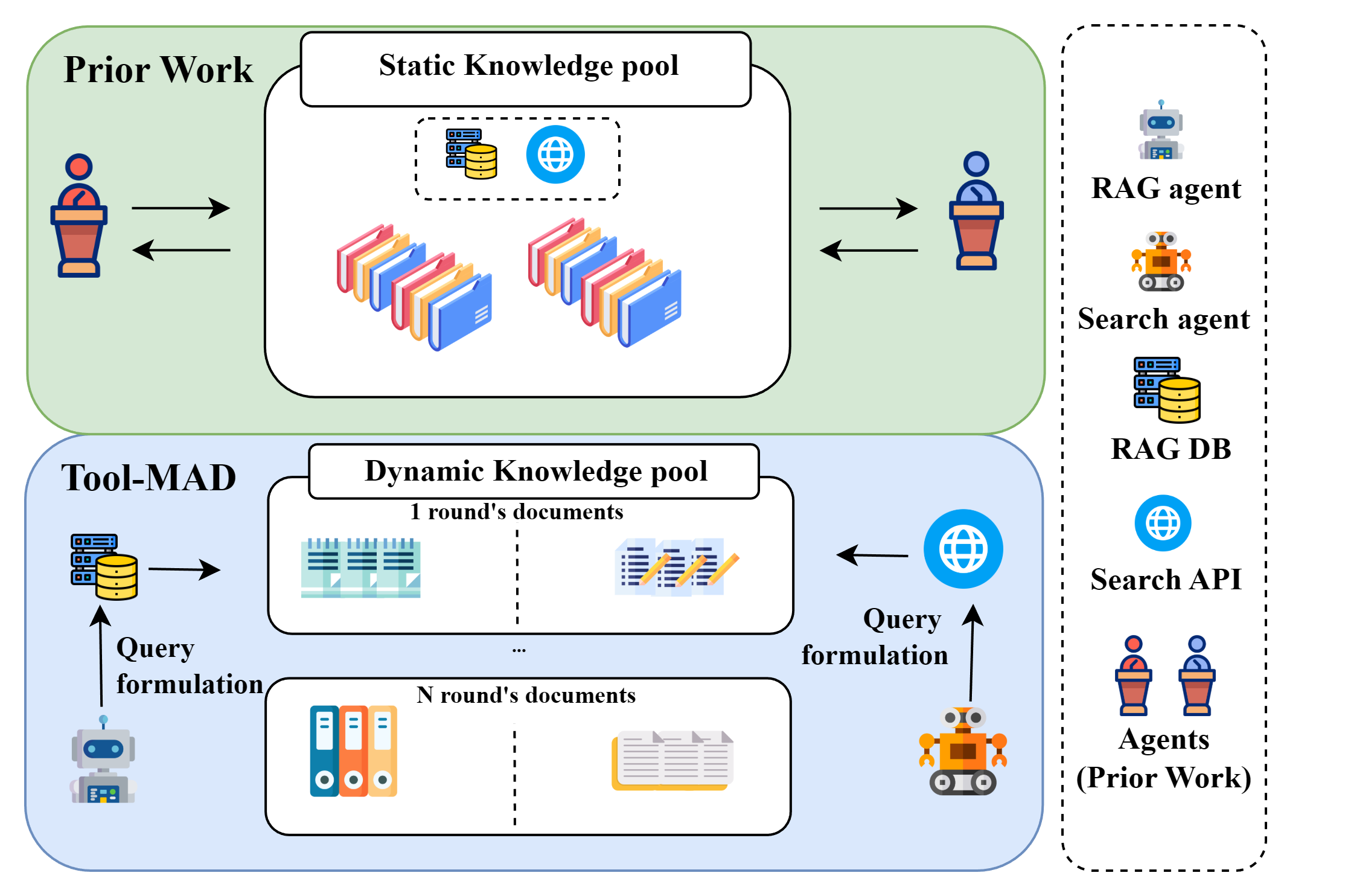}
\caption{Comparison of MAD~\cite{liang-etal-2024-encouraging}, MADKE~\cite{wang2025learning} and the proposed Tool-MAD: Unlike MAD and MADKE, which relies on a fixed document pool for retrieval, Tool-MAD dynamically retrieves external documents at each round through query formulation.}
\label{fig1}
\end{figure}

In this paper, we introduce Tool-MAD, a dynamic multi-agent debate framework designed to enhance factual verification by enabling agents to leverage heterogeneous external tools such as search APIs and Retrieval-Augmented Generation (RAG) modules, thereby promoting diverse perspectives~\cite{lewis2020retrieval}. Unlike prior multi-agent debate frameworks that rely on static retrieval or internally stored knowledge within the model, Tool-MAD is designed to allow agents to dynamically and iteratively retrieve new evidence as the debate progresses. Specifically, after each debate round, agents update their queries based on previous arguments exchanged during the discussion, enabling adaptive retrieval of relevant documents. This iterative knowledge acquisition process significantly enhances the adaptability of the framework, reduces the likelihood of hallucinations, and contributes to improving the factual reliability of the final decision. This mechanism is illustrated in Figure~\ref{fig1}, which compares MAD, MADKE and Tool-MAD.

In addition, we evaluate the responses generated in each debate round using two independent metrics from the RAGAS framework~\cite{es2024ragas}: faithfulness, which measures how well an agent’s claim is grounded in the retrieved evidence, and answer relevance, which assesses how directly the response addresses the original question. As these two metrics capture complementary aspects of response quality, we collectively refer to them as the stability score, which serves as an auxiliary signal for the Judge agent to assess the factual consistency and trustworthiness of each response.

We conduct extensive experiments across four fact verification benchmark datasets to evaluate the effectiveness of Tool-MAD. The results show that Tool-MAD consistently outperforms competitive multi-agent debate framworks such as MAD \cite{liang-etal-2024-encouraging} and MADKE \cite{wang2025learning}, achieving performance improvements of up to 35.5 $\%$ and 5.5 $\%$, respectively. Tool-MAD further demonstrates its flexibility in medical QA settings, maintaining robust performance under different retrieval tools and corpus configurations.

The main contributions of this paper can be summarized as follows:
\begin{itemize}
\item We propose Tool-MAD, a novel multi-agent debate framework that empowers agents to verify factual claims adaptively by leveraging a diverse set of external tools, including real-time search APIs and Retrieval-Augmented Generation (RAG) modules.
        
\item We introduce an adaptive query formulation mechanism, enabling agents to iteratively refine their evidence retrieval based on evolving debate contexts and previous arguments, leading to more informed and reliable judgments.
    
\item We incorporate faithfulness and answer relevance as a stability score in Tool-MAD to assess how well responses align with evidence and address the original question. This helps detect hallucinations and guide final decisions.

\item We comprehensively evaluate Tool-MAD on four benchmark datasets for fact verification, as well as two additional datasets for medical QA tasks, consistently surpassing competitive multi-agent debate baselines.
\end{itemize}

\section{Related Works}

\textbf{Multi-Agent Debate.}
Multi-agent debate frameworks have emerged as a promising direction for strengthening LLM reasoning by leveraging adversarial or collaborative interactions between agents. Early theoretical foundations stem from Minsky’s “society of minds’’ theory~\cite{minsky1986society}, which views intelligence as the emergent result of multiple interacting subsystems. This idea has resurfaced in modern LLM settings, where multiple agents exchange arguments, challenge assumptions, and refine intermediate reasoning.

Du et al.~\cite{du2023improving} demonstrated that multi-round debate enables LLMs to correct each other’s errors and improve logical consistency. Liang et al.~\cite{liang-etal-2024-encouraging} introduced a tit-for-tat debate structure to combat the \textit{Degeneration of Thought} problem identified in self-reflection methods~\cite{NEURIPS2023_1b44b878}. Other works have explored more complex multi-agent coordination strategies, such as majority voting, argument diversification, or structured deliberation trees. R\textsc{e}C\textsc{oncile}~\cite{chen2023reconcile} positions agents as participants in a roundtable discussion, integrating confidence-weighted consensus to avoid dominance by a single agent.

Beyond general reasoning, multi-agent frameworks have also been applied in specialized settings such as planning, safety evaluation, and code generation. For instance, debate-based self-correction mechanisms have been explored in mathematical reasoning, where agents critique intermediate steps, and in safety-alignment contexts, where disagreement is used to uncover unsafe model ~\cite{du2023improving}. These efforts show that debate can create richer reasoning traces, but they still rely heavily on internal knowledge.

MADKE~\cite{wang2025learning} introduced the idea of knowledge-enhanced debate by injecting externally retrieved documents before discussion; however, the evidence pool remains unchanged throughout the debate process. Similarly, debate-enhanced RAG systems retrieve supporting evidence prior to deliberation and then freeze the document set. While these systems show empirical gains, the fixed-evidence assumption limits adaptability when new arguments emerge during the debate.

\textbf{Fact Verification and Factuality Evaluation.}
Ensuring the factual correctness of LLM outputs is a long-standing challenge~\cite{li2024dawn, chen2023hallucination}. Early fact verification approaches relied on supervised classification models over structured evidence sources such as Wikipedia or news corpora. With the rise of generative models, research has shifted toward grounding generation in external corpora through retrieval-augmented generation (RAG)~\cite{lewis2020retrieval}. RAG-based fact-checkers~\cite{zhao2023felm} and hybrid evidence selection frameworks have shown that integrating retrieval significantly reduces hallucination in open-domain QA and claim verification.

Fact verification evaluation has similarly diversified. Works such as FEVER and its successors introduced label-based evaluation pipelines. More recent frameworks such as FactScore~\cite{min2023factscore} attempt to quantify factuality at the claim level by decomposing model outputs. Consistency-based factuality approaches, including self-consistency~\cite{wang2022self} and re-asking~\cite{manakul2023selfcheckgpt}, evaluate factual correctness by measuring the stability of model outputs rather than relying solely on external grounding.

RAGAS~\cite{es2024ragas} bridged retrieval-based reasoning and factuality evaluation by introducing two complementary metrics: \textit{faithfulness}, measuring whether claims match retrieved evidence, and \textit{answer relevance}, measuring whether responses address the question directly. Several subsequent works have adopted faithfulness for hallucination detection in RAG systems or used relevance-based filtering to identify unusable generations. However, these metrics have predominantly been applied as post-hoc evaluators for fully generated outputs, not as real-time signals that influence multi-step reasoning.

Furthermore, no prior work integrates metric-guided evaluation into a multi-agent debate structure, nor uses factuality signals to modulate argument selection, judge decisions, or evidence refinement across rounds. Tool-MAD incorporates these signals internally, using faithfulness and answer relevance as round-level stability indicators, which is distinct from existing factuality scoring frameworks.

\textbf{Tool-Augmented and Retrieval-Augmented Agents.}
A growing line of research explores augmenting LLMs with external tools, enabling them to perform tasks requiring specialized knowledge or computation. Toolformer~\cite{NEURIPS2023_d842425e} enables models to learn API-calling behaviors, while HuggingGPT~\cite{shen2023hugginggpt} and GEAR~\cite{lu2023gear} treat the LLM as a coordinator for heterogeneous models or systems. Tool-use frameworks have since evolved into modular agent pipelines that combine information extraction, symbolic reasoning, multi-step planning, or domain-specific simulators.

In scientific and professional domains, tool-augmented agents such as ChemCrow~\cite{bran2023chemcrow} and biomedical retrieval agents~\cite{gao2024empowering} demonstrate that domain-specific tools can dramatically improve reasoning fidelity. Retrieval modules—including dense retrievers, hybrid rankers, and cross-encoder rerankers—enable models to ground their reasoning in curated evidence.

However, existing tool-augmented systems overwhelmingly adopt a single-agent perspective and use tools in a one-shot or sequential fashion. They do not exploit multi-agent interaction as a mechanism for tool selection, evidence diversification, or iterative grounding. Likewise, retrieval systems typically perform a single retrieval step at the beginning of a task, without adapting to evolving reasoning trajectories or emerging counterarguments. As a result, these systems often fail to revisit or refine earlier tool calls when new uncertainties arise, leading to brittle reasoning paths. Moreover, the lack of interaction between agents prevents the system from leveraging disagreement or complementary viewpoints to guide more targeted retrieval or tool use.

\begin{figure*}[t]
\centering
\includegraphics[width=1.0\textwidth]{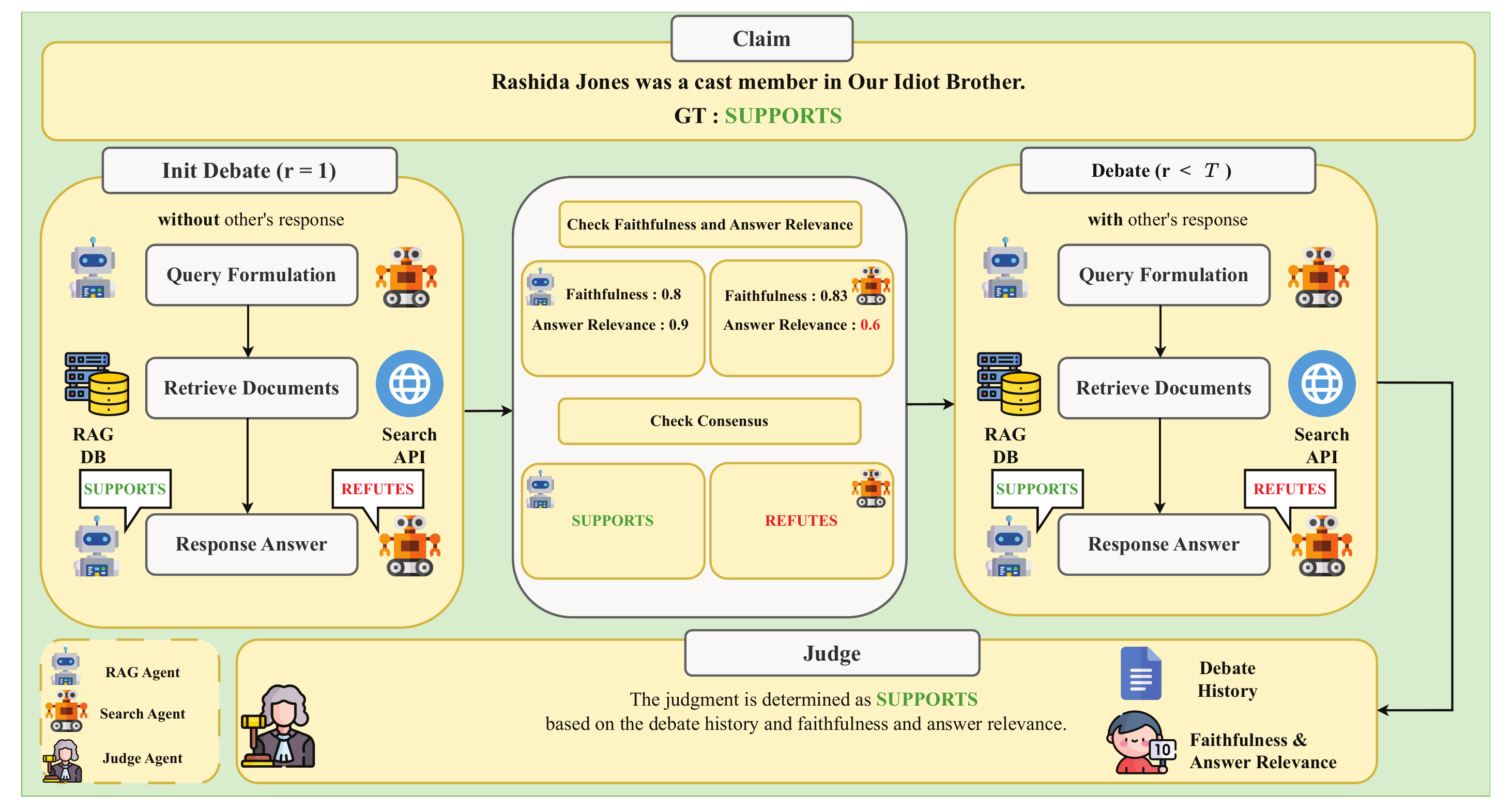}
\caption{Given a claim, two agents (RAG and Search) engage in multi-round debates, where $r$ denotes the current round and $T$ is the predefined round threshold. If no consensus is reached or the stability score falls below the threshold by round $T$, the Judge Agent issues a final verdict based on the debate history and the stability score.}
\label{fig2}
\end{figure*}

\section{Tool-MAD}
In this section, we introduce Tool-MAD, a novel multi-agent debate framework designed to enhance the factual reliability of LLMs for claim verification tasks. Unlike previous methods that rely primarily on static evidence or single-agent reasoning, Tool-MAD incorporates iterative retrieval of external evidence and dynamic interactions among multiple specialized agents. By repeatedly updating evidence and challenging one another’s conclusions, the system progressively refines its reasoning, mitigates hallucinations, and promotes more reliable consensus formation.

\subsection{External Tools}
To enable dynamic, context-aware fact verification, Tool-MAD equips agents with external retrieval tools that provide access to relevant and timely evidence during debates. Specifically, we integrate two complementary retrieval mechanisms: a RAG module leveraging a static corpus, and a live web Search API for real-time information access.

\textbf{Retrieval-Augmented Generation} We employ the RAG framework~\cite{guu2020retrieval} to augment agent reasoning with relevant documents retrieved from an embedded vector store. For this purpose, we use Milvus~\cite{2021milvus}, a scalable and efficient vector database optimized for high-dimensional corpus management. We index a corpus constructed from Wikipedia articles, enabling rapid semantic retrieval. At inference time, each query returns the top three most semantically relevant documents, providing agents with targeted supporting evidence.

\textbf{Search API} Complementing the static RAG corpus, Tool-MAD also incorporates a real-time Search API, enabling agents to access up-to-date information directly from the web. For this, we utilize the Tavily Search API~\footnote{https://tavily.com/}, known for its effective integration with language models. Similar to the RAG system, the Search API retrieves the three most relevant documents per query, ensuring comprehensive coverage of evolving knowledge demands during the debate.

\begin{algorithm}[H]
\caption{Tool-MAD Framework}
\label{alg:toolmad}
\begin{algorithmic}
\STATE \textbf{Input:} claim $c$, debater set $\mathcal{A}$, max rounds $T$, stability thresholds $St = (f, ar)$
\STATE \textbf{Output:} final decision answer

\STATE Initialize history $H \gets \emptyset$
\FOR{each $A \in \mathcal{A}$}
    \STATE $S^{A} \gets 0$
\ENDFOR

\FOR{$r = 1$ \textbf{to} $T$}

    \FOR{each agent $A \in \mathcal{A}$}
        \STATE Let $\bar A$ be the opponent of $A$

        \IF{$r = 1$}
            \STATE $q_A^{(r)} \gets \mathit{Query}(A, c)$
            \STATE $D_A^{(r)} \gets \mathit{Retrieve}(A, q_A^{(r)})$
            \STATE $a_A^{(r)} \gets \mathit{Respond}(A, D_A^{(r)}, c)$
        \ELSE
            \STATE $q_A^{(r)} \gets \mathit{Query}(A, c, a_{\bar A}^{(r-1)}, q_A^{(r-1)})$
            \STATE $D_A^{(r)} \gets \mathit{Retrieve}(A, q_A^{(r)})$
            \STATE $a_A^{(r)} \gets \mathit{Respond}(A, D_A^{(r)}, c, a_{\bar A}^{(r-1)})$
        \ENDIF

        \STATE $f_A^{(r)} \gets \mathit{faithfulness}(a_A^{(r)}, D_A^{(r)})$
        \STATE $ar_A^{(r)} \gets \mathit{answerRelevance}(c, a_A^{(r)})$
        \STATE $S^{A} \gets S^{A} + (f_A^{(r)},\; ar_A^{(r)})$
    \ENDFOR

    \STATE Append round-$r$ info to $H$

    \IF{$a_R = a_S$ \AND $S^R > St$ \AND $S^S > St$}
        \STATE \textbf{return} $a_R$
    \ENDIF
\ENDFOR

\STATE \textbf{return} $\mathit{Judge}(A_J, H, \{S^{A}\}_{A \in \mathcal{A}}, c)$

\end{algorithmic}
\end{algorithm}

\subsection{Debate Participants}

\textbf{Debater Agents} Tool-MAD involves two specialized debater agents: a RAG-based agent ($A_R$) that retrieves evidence from a static vector-based corpus, and a Search-based agent ($A_S$) that accesses live web documents via a search API. Both agents operate under a shared prompt template to ensure consistent behavior across rounds. In the first round, each agent independently generates a response based solely on the input claim. In subsequent rounds, agents refine their responses by incorporating the previous response from their opponent, enabling richer reasoning and exposure to diverse viewpoints.

\textbf{Judge Agent} If the two debater agents fail to reach consensus within a predefined maximum number of rounds, the final decision is made by a third agent, the Judge ($A_J$). The Judge determines the outcome based on three primary inputs: (1) the original claim, (2) the full debate history—including agent responses and retrieved evidence from each round, and (3) a stability score that assesses the factual consistency and relevance of each agent’s arguments. 

\subsection{Stability Score}

To quantitatively evaluate the reliability of each agent's response, Tool-MAD adopts two core metrics from the RAGAS framework~\cite{es2024ragas}: \textit{faithfulness} and \textit{answer relevance}. These two metrics jointly form the \textit{Stability Score} used throughout the debate.

\textbf{Faithfulness} measures how accurately an agent’s response reflects the content of the retrieved evidence. 
Each response $a_s(q)$ is decomposed into a set of factual statements:
\[
S(a_s(q)) = \{ s_1, s_2, \ldots, s_{|S|} \}.
\]
For each statement $s_i$, the LLM determines whether it can be inferred from the retrieved context $c(q)$. 
We define a verification function $v(s_i, c(q))$ such that $v(s_i, c(q)) = 1$ if $s_i$ is supported by $c(q)$, and $v(s_i, c(q)) = 0$ otherwise.
The faithfulness score $F$ is then computed as the proportion of supported statements:
\[
F = \frac{\sum_{i=1}^{|S|} v(s_i, c(q))}{|S|}.
\]
A high faithfulness score indicates that the response is well-grounded and factually consistent with the retrieved evidence.

\textbf{Answer relevance} assesses how directly the agent’s response addresses the original question. 
For the given answer $a_s(q)$, the LLM generates a set of $n$ potential questions:
\[
Q' = \{ q_1, q_2, \ldots, q_n \},
\]
where each $q_i$ is intended to represent a question that the answer could plausibly correspond to. 
We compute embeddings for the original question $q$ and each generated question $q_i$, and calculate their cosine similarity:
\[
sim(q, q_i) = \cos\big( \mathbf{e}(q), \mathbf{e}(q_i) \big).
\]
The answer relevance score is then obtained by averaging the similarities:
\[
AR = \frac{1}{n} \sum_{i=1}^{n} sim(q, q_i).
\]
A high relevance score suggests that the answer remains focused on the intended question, whereas a low score indicates incompleteness or topic drift.

Both metrics are computed at every debate round. If either faithfulness or answer relevance falls below a predefined threshold, the round is marked as inconclusive and the debate continues. 

\begin{figure*}
    \centering
    \includegraphics[width=0.9\linewidth]{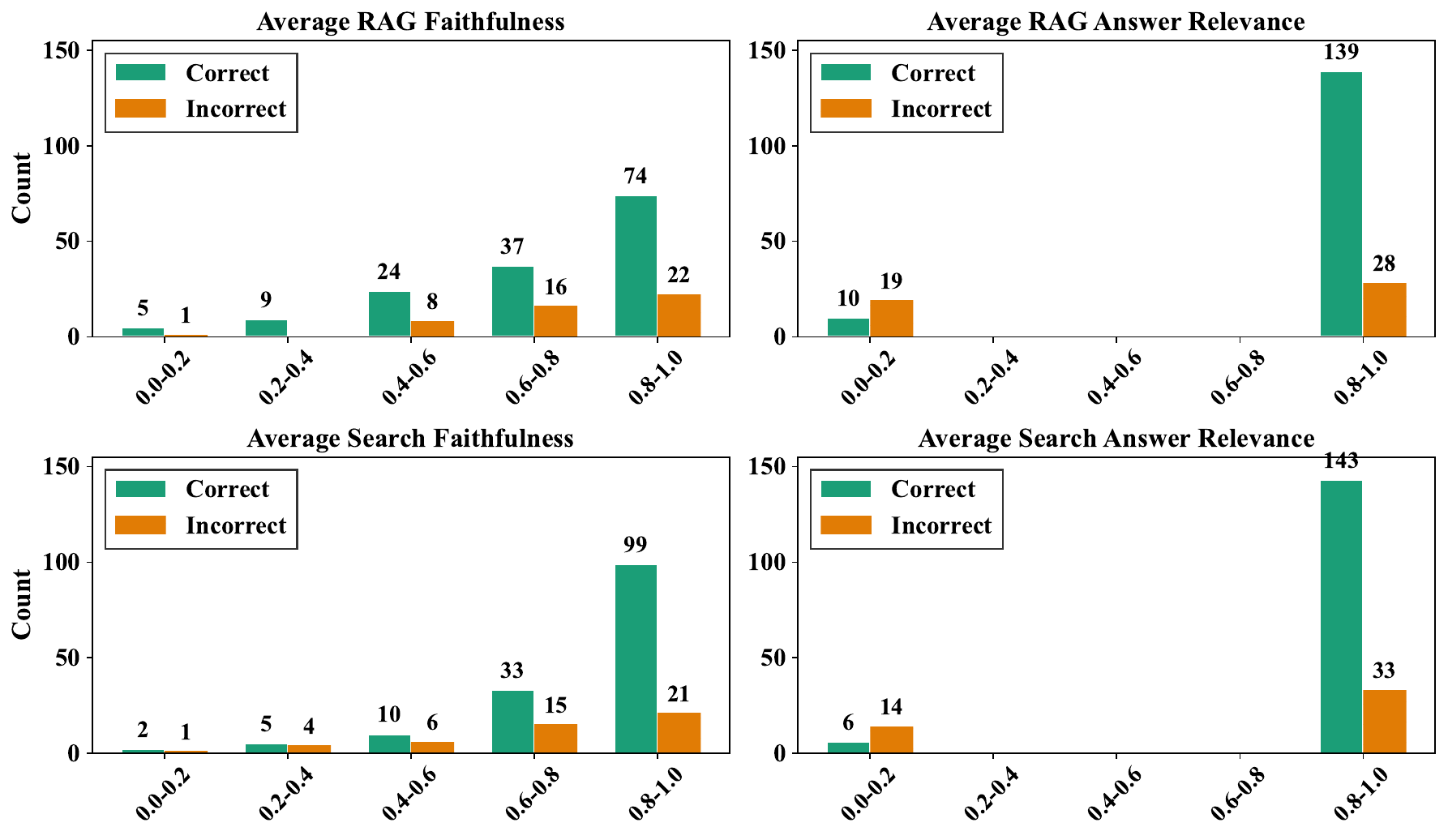}
    \caption{Empirical distributions of faithfulness and answer relevance for selecting stability-score thresholds. Answer relevance is concentrated above 0.8, while faithfulness shows a wider spread around 0.7–0.8. We therefore set thresholds to 0.7 (faithfulness) and 0.8 (answer relevance) to balance precision and efficiency}
    \label{threshold}
\end{figure*}

\textbf{Threshold of Stability Score}
To determine suitable stability-score thresholds, we analyzed the empirical distributions of faithfulness and answer relevance (Figure~\ref{threshold}). Answer relevance was strongly skewed toward high values, with most scores above 0.8, making 0.8 a natural cutoff that filters low-quality outputs without affecting the majority of valid ones. In contrast, faithfulness exhibited a wider spread, with many responses falling between 0.7 and 0.8; thus, using a stricter threshold would unnecessarily reject reasonable, evidence-aligned answers and prolong debates.

These observations reveal a key trade-off: overly strict thresholds increase precision but hurt efficiency, especially when models produce borderline yet acceptable outputs. Therefore, we adopt 0.7 for faithfulness and 0.8 for answer relevance, which empirically provides stable convergence while maintaining decision quality.

\begin{figure}[t]
\centering
\includegraphics[width=0.9\columnwidth]{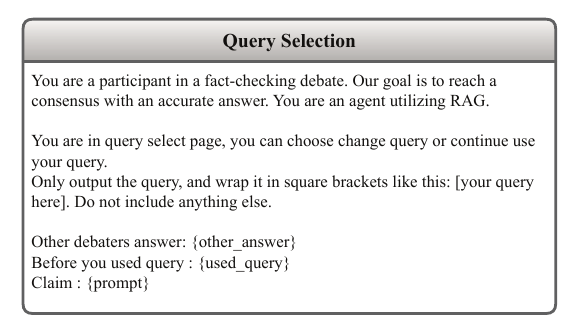} 
\caption{Prompt of query selection}
\label{prompt:fig4}
\end{figure}

\begin{figure}[t]
\centering
\includegraphics[width=0.9\columnwidth]{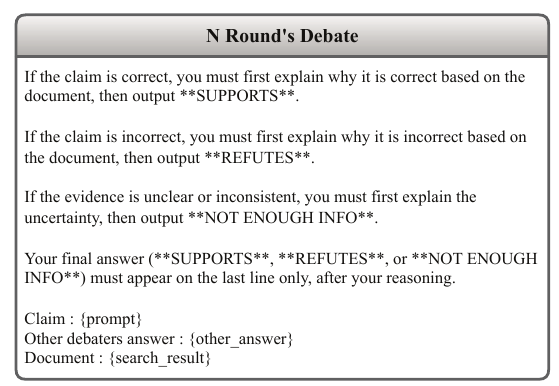} 
\caption{Prompt of round debate}
\label{prompt:fig5}
\end{figure}

\begin{figure}[t]
\centering
\includegraphics[width=0.9\columnwidth]{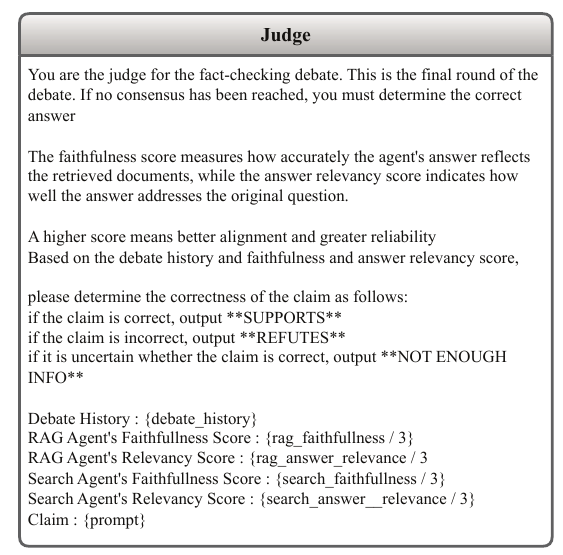}
\caption{Prompt of judge}
\label{prompt:fig6}
\end{figure}

\subsection{Tool-MAD Procedure}

Tool-MAD proceeds in multiple rounds of interaction between two debater agents, each equipped with distinct external tools. Let $c$ denote the input claim, and $r \in \{1, \ldots, T\}$ be the current debate round, where $T$ is the maximum allowed number of rounds. The framework involves three agents: a retrieval-based debater $A_R$, a search-based debater $A_S$, and a Judge agent $A_J$. Each debater generates arguments based on its retrieved evidence, and the responses in each round are quantitatively evaluated using faithfulness and answer relevance scores. If either agent's response fails to meet predefined thresholds for these scores, or if no consensus is reached, the debate proceeds to the next round. If consensus is not achieved by the final round $T$, the Judge agent produces a final verdict based on the accumulated dialogue and retrieved evidence. The overall framework overview is presented in Figure~\ref{fig2}, and the complete algorithmic details are provided in the algorithm ~\ref{alg:toolmad}.

\textbf{Initialization Round ($r = 1$).}  
In the first round, each debater independently constructs an initial query based solely on the input claim $c$, without reference to the opponent's argument. This query is submitted to the agent’s designated retrieval tool, which returns a set of top-$k$ relevant documents. The agent then composes its response by reasoning over the claim and the retrieved evidence. For agent $A \in \{A_R, A_S\}$, the process is formally defined as:

\begin{equation}
q_A^1 = Query(A, c)
\end{equation}

\begin{equation}
D_A^1 = Retrieve(A, q_A^1)
\end{equation}

\begin{equation}
a_A^1 = Respond(A, D_A^1, c)
\end{equation}

where $q_A^1$ is the retrieval query, $D_A^1$ is the set of retrieved documents, and $a_A^1$ is the generated response for round 1.

\textbf{Debate Rounds ($r > 1$).}  
In each subsequent round, agents refine their queries and responses by incorporating the opponent’s previous answer. This enables dynamic evidence updates and encourages more robust reasoning. Each round consists of three steps: query formulation, evidence retrieval, and response generation. Specifically:

\begin{equation}
    q_A^r = Query(A, c, a_{\bar{A}}^{r-1})
\end{equation}
\begin{equation}
    D_A^r = Retrieve(A,q_A^r)
\end{equation}
\begin{equation}
    a_A^r = Respond(A,D_A^r,c,a_{\bar{A}}^{r-1})
\end{equation}
where $a_{\bar{A}}^{r-1}$ is the previous response from the opposing agent $\bar{A}$.

If both agents produce the same response in any round, the debate terminates early with a consensus, considering predefined thresholds of the stability score. Otherwise, the current round is appended to the debate history. If consensus is not reached or the responses fall below the thresholds for faithfulness or answer relevance, the debate continues until round $T$. The Judge agent $A_J$ then determines the final outcome.

\textbf{Judge Decision(triggered only when the debate reaches the final round)}  
If the agents fail to reach a consensus by the final round $T$, the Judge agent $A_J$ makes the final decision based on the original claim $c$, the complete debate history $H$, and the stability score $S$. Here, $S$ is composed of the average faithfulness and answer relevance scores across all rounds, capturing the overall quality of the agents' responses. This decision process is formalized as:

\begin{equation}
    final\,answer = Judge(A_J, H, S, c)
\end{equation}

The LLM-based Judge agent receives the stability score $S$ and debate history $H$ as part of its prompt and makes the final decision by jointly considering both.

For reproducibility, we provide the full prompts used for the debate round, query selection, and judge components in Figures ~\ref{prompt:fig4},  ~\ref{prompt:fig5}, and ~\ref{prompt:fig6}. Note that the prompt for the initial round corresponds to the same template with the opponent’s response removed.

\begin{table}[t]
    \centering
    \caption{Datasets used in Tool-MAD, categorized by task type and corresponding references.}
    \begin{tabular}{llllll}
        \hline
        \textbf{Dataset} &
        \textbf{Task} & \\
        \hline
        \textbf{FEVER \cite{thorne2018fever}} & Fact Verification \\
        \textbf{FEVEROUS \cite{aly2021feverous}}& Fact Verification\\
        \textbf{\textsc{FaVIQ} \cite{park2021faviq}} & Fact Verification \\
        \textbf{\textsc{AV}\textsc{eri}\textsc{TeC} \cite{schlichtkrull2023averitec}} & Fact Verification \\
        \hline
        \textbf{\textsc{MedQA} \cite{jin2021disease}} & Medical \\
        \textbf{\textsc{PubMedQA} \cite{jin2019pubmedqa}} & Medical \\
        \hline
    \end{tabular}
    \label{tab:dataset}
\end{table}

\begin{table*}[t]
    \centering
    \caption{Structural comparison of baseline reasoning frameworks.}
    \begin{tabular}{lcccccc}
        \hline
        \textbf{Model} & \textbf{Reasoning} & \textbf{Multi-Agent} & \textbf{Retrieval} & \textbf{Query Formulation}\\
        \hline
        CoT (Zero-shot)        & O & X & X & X \\
        Single-Agent (ReAct)   & O & X & O & O  \\
        MAD                    & O & O & X & X  \\
        MADKE                  & O & O & O & X \\
        \textbf{Tool-MAD}      & O & O & O & O \\
        \hline
    \end{tabular}
    \label{table:structure_comparison}
\end{table*}

\section{Experiments}
 We evaluate the effectiveness of a Tool-MAD framework that leverages external tools in performing the fact verification task. Unless specified otherwise, all experiments were conducted on 200 randomly sampled instances per dataset, evaluated using an Exact Match (EM) criterion, where a prediction is considered correct if it exactly matches the ground truth labels.
 
\subsection{Datasets}
We evaluate our method across a wide range of tasks to assess both factual accuracy and cross-task flexibility. For fact verification, we utilize four widely used benchmark datasets: FEVER \cite{thorne2018fever} and FEVEROUS \cite{aly2021feverous}, which are based on Wikipedia-derived claims, as well as F\textsc{a}VIQ \cite{park2021faviq} and \textsc{AV}\textsc{eri}\textsc{TeC} \cite{schlichtkrull2023averitec}, which include claims from real world contexts. These datasets collectively span diverse domains and claim structures, allowing for a robust assessment of factual verification capabilities. To evaluate the flexibility of our framework beyond fact verification, we additionally include medical QA datasets such as M\textsc{ed}QA \cite{jin2021disease} and PubMedQA \cite{jin2019pubmedqa}, which focus on domain-specific clinical and biomedical reasoning. A detailed summary of all datasets, including scale, domain, and task type, is provided in Table~\ref{tab:dataset}.

\subsection{Models} We employ GPT-4o-mini \cite{4oapi} and Llama-3.3-70B-Instruct-Turbo \cite{grattafiori2024llama} as backbone models in our Tool-MAD framework. We also utilize GPT-4o \cite{4oapi}, a larger variant of GPT-4o-mini, for performance comparison. To further assess the effectiveness of the proposed framwork, we conduct a performance comparison with DeepseekR1 \cite{guo2025deepseek}, a representative reasoning based model. 

\begin{table*}[t]
    \centering
    \caption{Main results on four fact verification datasets (FEVER, FEVEROUS, FAVI-Q, and AVeriTeC). Tool-MAD consistently improves performance across both proprietary (e.g., GPT-4o, DeepSeekR1) and open-source (e.g., LLaMA-3.3-70B) models, outperforming other multi-agent debate frameworks. Average scores are reported for overall comparison. \textbf{Bold} indicates the highest Exact Match score within each base model group (above: GPT-4 variants and DeepSeekR1; below: LLaMA).}

    \setlength{\tabcolsep}{14pt}

    \begin{tabular}{lccccc}
        \hline
        \textbf{Model} &
        \textbf{FEVER} &
        \textbf{FEVEROUS} &
        \textbf{F\textsc{a}VIQ} &
        \textbf{\textsc{AV}\textsc{eri}\textsc{TeC}} &
        \textbf{Average}\\
        \hline
        DeepseekR1 & 71.0 & 67.5 & 77.0 & 49.0 & 66.1 \\
        GPT-4o & 69.5 & 54.5 & 68.0 & 43.5 & 58.9 \\
        GPT-4o-mini & 62.0 & 37.0 & 56.0 & 33.0 & 47.0 \\
        + CoT(Zero-shot) & 66.5 & 31.0 & 67.0 & 33.5 & 49.5 \\
        + Single Agent(ReAct) & 62.0 & 24.0 & 66.0 & 24.0 & 44.0\\
        + MAD & 71.0 & 36.5 & 68.0 & 36.0 & 52.9\\
        + MADKE & 72.0 & 66.0 & 75.5 & 58.5 & 68.0 \\
        + Tool-MAD & \textbf{73.0} & \textbf{71.5} & \textbf{77.5} & \textbf{62.0} & \textbf{71.0}\\
        \hline
        Llama-3.3-70B(Inst) & 69.5 & 49.0 & 64.0 & 51.0 & 58.4\\
        + CoT(Zero-shot) & 71.0 & 51.0 & 68.5 & 32.0 & 55.6\\
        + Single Agent(ReAct) & 73.5 & 51.0 & 65.5 & 35.5 & 56.4\\
        + MAD & 54.0 & 31.5 & 58.5 & 39.5 & 45.9\\
        + MADKE & 62.5 & 61.0 & 71.5 & 31.0 & 56.5\\
        + Tool-MAD & \textbf{74.0} & \textbf{77.0} & \textbf{78.5} & \textbf{66.5} & \textbf{74.0}\\
        \hline
    \end{tabular}

    \label{table:table1}
\end{table*}

\begin{table}[t]
\centering
\caption{Accuracy and 95\% confidence intervals across datasets. (Backbone model : GPT-4o-mini)}
\begin{tabular}{lcc}
\hline
\textbf{Dataset} & \textbf{Acc.} & \textbf{95\% C.I} \\
\hline
FEVER    & 0.73 & (0.67, 0.79) \\
FEVEROUS & 0.72 & (0.66, 0.78) \\
FAVIQ    & 0.78 & (0.72, 0.83) \\
AVeriTeC & 0.62 & (0.56, 0.69) \\
\hline
\end{tabular}
\label{table:2}
\end{table}

\textbf{Baseline Model} We evaluate Tool-MAD by comparing it with single agent reasoning and multi-agent debate baselines. 
A brief summary is provided below:
    \begin{itemize}
        \item \textbf{Zero-shot CoT} \cite{kojima2022large} : Zero-shot CoT is to induce reasoning with the prompt “Let’s think step by step".
        \item \textbf{Single Agent} \cite{yao2023react} : The single-agent baseline is based on the ReAct framework, using RAG for external knowledge retrieval. ReAct enables iterative reasoning via tool use and feedback.
        \item \textbf{MAD} \cite{liang-etal-2024-encouraging} : MAD is a framework that uses an interactive “tit for tat" debate structure, motivated by the Degeneration of Thought observed in single agents, even when self-reflection is applied.
        \item \textbf{MADKE} \cite{wang2025learning} : MADKE addresses the limitations of traditional MAD methods that rely solely on internal knowledge by incorporating a static evidence pool retrieved prior to the debate.
    \end{itemize}

Table~\ref{table:structure_comparison} provides a structural comparison of these baselines, highlighting their differences across four key dimensions: intrinsic reasoning capability, multi-agent interaction, retrieval usage, and query formulation. As shown in the table, Tool-MAD is the only framework that simultaneously supports all four components by integrating structured debate, iterative retrieval, and adaptive query rewriting. In contrast, prior baselines typically cover only a subset of these capabilities, resulting in more limited overall functionality.

\begin{table}[t]
  \centering
  \caption{Exact Match (EM) scores of different multi-agent debate frameworks on MedQA and PubMedQA. \textbf{Bold} indicates the highest score.}

  \setlength{\tabcolsep}{12pt}

  \begin{tabular}{lcc}
    \hline
    \textbf{Model} & \textbf{MedQA} & \textbf{PubMedQA} \\
    \hline
    MAD          & 58.0           & 22.5              \\
    MADKE        & 74.0           & 21.5              \\
    Tool-MAD     & \textbf{77.0}  & \textbf{29.0}     \\
    \hline
  \end{tabular}

  \label{tab:medqa_pubmedqa_em}
\end{table}

\begin{table*}[t]
    \centering
    \caption{Performance comparison of different agent combinations on FEVER, FEVEROUS, \textsc{FaVIQ} and \textsc{AV}\textsc{eri}\textsc{TeC} datasets. Tool-MAD achieves the best overall results across combinations of \textsc{RAG}, \textsc{Search}, and \textsc{Vanila} agents. \textbf{Bold} indicates the highest Exact Match score.}

    \setlength{\tabcolsep}{14pt}

    \begin{tabular}{lcccc}
        \hline
        \textbf{Model} &
        \textbf{FEVER} &
        \textbf{FEVEROUS} &
        \textbf{F\textsc{a}VIQ} & 
        \textbf{\textsc{AV}\textsc{eri}\textsc{TeC}}\\
        \hline
        \textsc{Vanila + Vanila} & 62.0 & 40.0 & 60.0 & 45.0\\
        \textsc{RAG} + \textsc{Vanila}  & 65.5 & 57.5 & 63.5 & 58.0\\
        \textsc{Search} + \textsc{Vanila}  & 60.5 & 43.0 & 63.5 & 53.5\\
        \textsc{RAG} + \textsc{RAG} & 67.5 & 60.5 & 68.5 & 56.5 \\
        \textsc{Search} + \textsc{Search} & 67.0 & 60.5 & 68.5 & 55.5 \\
        Tool-MAD (\textsc{RAG} + \textsc{Search}) & \textbf{73.0} & \textbf{71.5} & \textbf{77.5} & \textbf{62.0} \\
        \hline
    \end{tabular}

    \label{combination}
\end{table*}

\subsection{Fact Verification Experiments}
\label{sec:4-3}
We evaluate multi-agent debate models on four fact verification benchmark datasets. Table~\ref{table:table1} presents the experimental results. Tool-MAD, using the lightweight GPT-4o-mini as its backbone, outperforms the more powerful GPT-4o across all evaluated benchmark datasets. Notably, it shows a 18.5$\%$ improvement on \textsc{AV}\textsc{eri}\textsc{TeC}, the dataset with the highest label complexity, demonstrating the robustness of our framework under more challenging verification settings. Compared to DeepSeekR1, Tool-MAD consistently achieves higher accuracy across all tasks, with improvements of up to $13\%$. Tool-MAD achieves comparable performance with the open-source Llama-3.3-70B backbone, confirming its robustness across model backbones.

Tool-MAD outperforms other multi-agent debate frameworks, achieving up to 35.0$\%$ and 5.5$\%$ improvements over MAD and MADKE, respectively, with average gains of 18.1$\%$ and 3.0$\%$. These results demonstrate that Tool-MAD delivers superior performance in fact verification tasks compared to existing multi-agent frameworks, and further highlight its architectural advantage over MADKE, which also leverages external knowledge.

To better understand the observed performance gap, we analyze the limitations of each baseline system. The Single Agent, despite leveraging the ReAct framework~\cite{yao2023react} and external tools, frequently generates incorrect answers due to the retrieval of irrelevant or incomplete documents. MAD~\cite{liang-etal-2024-encouraging}improves upon this through multi-agent debate, but still struggles when evidence is insufficient, often defaulting to a generic “Not Enough Info” response. MADKE~\cite{wang2025learning} achieves relatively stable performance by relying on a static evidence pool retrieved prior to the debate, yet fails to generalize well on more complex datasets such as \textsc{AV}\textsc{eri}\textsc{TeC} and FEVEROUS, where dynamic adaptation is required. These results highlight the architectural advantage of Tool-MAD, which enables agents to iteratively retrieve new and contextually relevant evidence throughout the debate, leading to improved factual reasoning. 

Additionally, we report 95\% bootstrap confidence intervals for Tool-MAD using GPT-4o-mini as the backbone. Due to the computational cost of multi-round multi-agent debate—where each query triggers multiple LLM inferences—conducting large-scale repeated trials is prohibitively expensive. The reported confidence intervals (Table II) therefore serve to supplement the robustness of our findings by quantifying the variability of the results under resampling. The corresponding results are presented in Table ~\ref{table:2}.

\begin{figure*}
    \centering
    \includegraphics[width=0.75\linewidth]{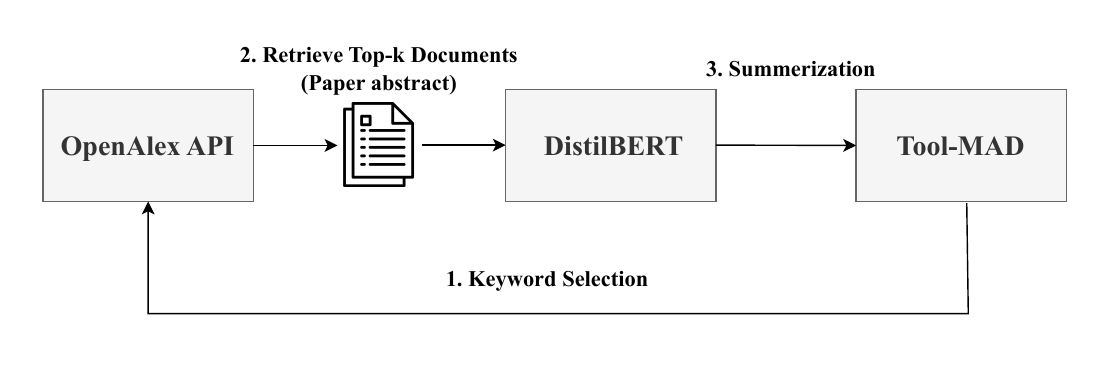}
    \caption{Workflow diagram of the paper retrieval API used in the PubMedQA experiment}
    \label{fig7}
\end{figure*}

\subsection{Flexibility}
This experiment focuses on evaluating whether Tool-MAD can effectively adapt and maintain consistent performance when external tools are changed or applied to different domains. For comparison, we conduct flexibility experiments using the frameworks that showed competitive performance in Fact Verification Experiments section, in place of single-agent baselines that demonstrated relatively lower performance. We evaluate our framework on two medical QA datasets: M\textsc{ed}QA, which focuses on clinical multiple-choice questions, and PubMedQA, which targets biomedical fact verification. To test the framework’s extensibility across tools, we configure the M\textsc{ed}QA experiment using a PubMed-based RAG corpus, while the PubMedQA setting replaces the standard search API with OpenAlex \cite{priem2022openalex}, an open scholarly database. 

In the PubMedQA pipeline, the agent extracts keywords from each query, retrieves three relevant abstracts, summarizes them using BERT \cite{devlin2019bert}, and incorporates the summaries into the debate. Detailed information on the PubMedQA workflow is provided in the figure ~\ref{fig7}. 

As shown in Table \ref{tab:medqa_pubmedqa_em}, Tool-MAD outperforms other multi-agent debate frameworks on both datasets. Notably, while MADKE performs worse than MAD on PubMedQA despite incorporating external knowledge, Tool-MAD achieves strong results and maintains consistent performance even when the underlying tools are changed.

Beyond these quantitative results, the strong performance of Tool-MAD on both M\textsc{ed}QA and PubMedQA can be attributed to the complementary nature of its heterogeneous retrieval sources. Clinical QA tasks often require both up-to-date evidence, such as newly published studies or evolving treatment guidelines, and structured biomedical knowledge grounded in established definitions or mechanistic explanations. Search-based retrieval is well suited for capturing the former, whereas RAG-based retrieval excels at providing the latter. By integrating both types of information within the debate, Tool-MAD enables agents to form more comprehensive and reliable arguments compared to methods that rely on a single evidence modality.

Another noteworthy aspect is that Tool-MAD maintains stable performance despite the domain shift from Wikipedia-style fact verification to biomedical question answering. This robustness arises from the framework’s iterative retrieval mechanism, which allows agents to refine their queries and adjust their evidence sources as domain-specific reasoning challenges emerge. The ability to dynamically cross-validate retrieved evidence across rounds helps mitigate errors introduced by unfamiliar terminology or dataset-specific retrieval noise.

Finally, the debate structure used in Tool-MAD naturally aligns with multi-source reasoning patterns observed in clinical decision-support workflows. In practice, healthcare professionals often synthesize information from both standardized guidelines and recent studies, and Tool-MAD mirrors this process by enabling different agents to contribute distinct yet complementary evidence. The use of faithfulness and answer relevance as stability indicators further helps filter out unsupported or clinically unreliable responses, providing an additional layer of safety when operating in high-stakes medical domains.

\begin{figure}[t]
\centering
\includegraphics[width=0.8\columnwidth]{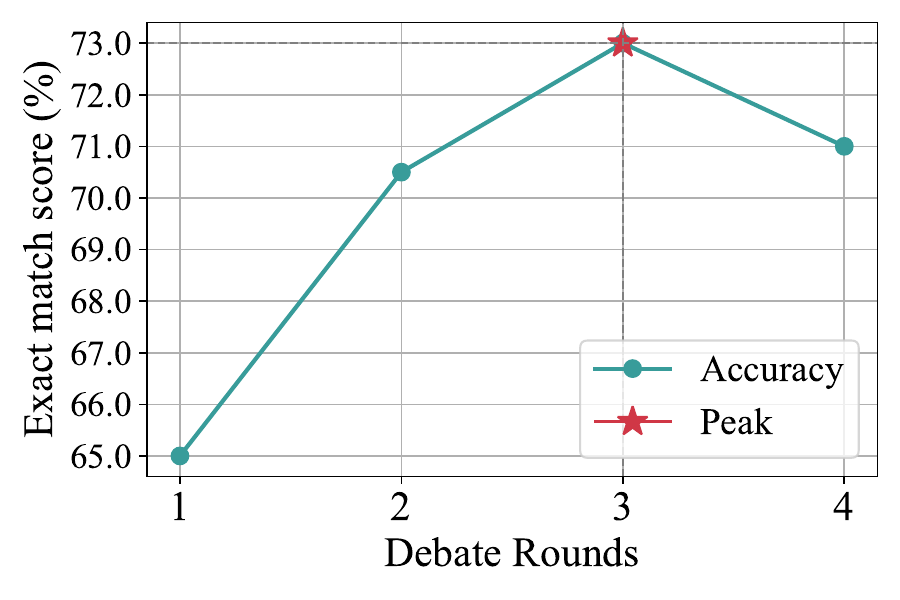}
\caption{Exact Match performance of Tool-MAD on
the FEVER dataset across different debate rounds.}
\label{fig3}
\end{figure}

\subsection{Ablation Study}

\textbf{Combination of Agents} To evaluate the impact of different agent configurations, we compare their performance across the FEVER, FEVEROUS, and F\textsc{a}VIQ benchmark datasets. All agents follow the same Tool-MAD procedure, with the only variation being the external tools they employ. Specifically, we define three types of agents: a base agent without any external tools (\textsc{Vanilla}), an agent using retrieval-augmented generation (\textsc{RAG}), and an agent equipped with a web-based search API (\textsc{Search}).

The results are presented in Table~\ref{combination}. As shown in the table, configurations that incorporate external tools consistently outperform those without tools, indicating that tool integration improves performance in fact verification tasks. Among all agent combinations tested, the original Tool-MAD setup (\textsc{RAG} + \textsc{Search}) achieved the highest average accuracy, demonstrating its robustness and effectiveness.

\begin{figure}[t]
\centering
\includegraphics[width=0.9\columnwidth]{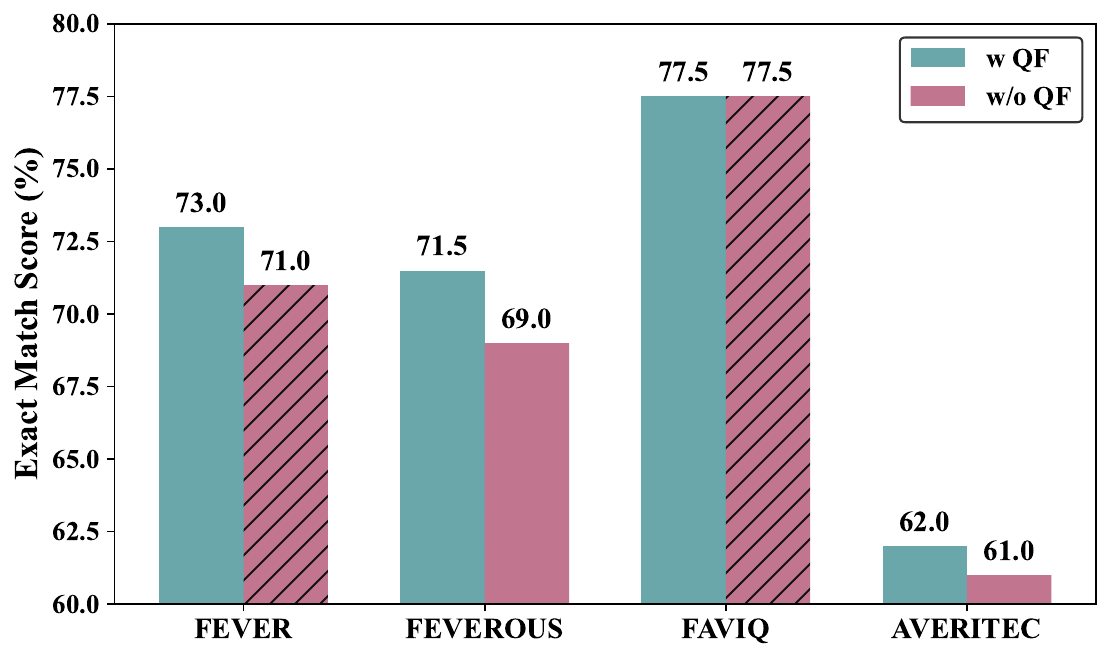}
\caption{Exact Match performance comparison with
and without query formulation (QF) across four benchmark
datasets.}
\label{fig4}
\end{figure}

In particular, configurations like \textsc{RAG}+\textsc{RAG} or \textsc{Search}+\textsc{Search} outperformed the \textsc{Vanilla} setup, but still fell short of the hybrid configuration (\textsc{RAG}+\textsc{Search}). This suggests that relying on a single type of tool can lead to redundant retrievals and limit the diversity of evidence, thereby constraining overall performance improvements.

To further interpret these results, we note that the hybrid configuration benefits from the complementary characteristics of the two external tools. RAG-based retrieval provides semantically aligned, context-rich documents, while search-based retrieval offers broader coverage and access to rapidly updated information. These heterogeneous evidence sources allow the agents to explore distinct perspectives during the debate, reducing shared blind spots and producing more reliable conclusions. In contrast, homogeneous configurations tend to retrieve overlapping content, which limits the diversity of arguments and constrains potential performance gains.

\vspace{0.2cm}

\begin{figure}[t]
\centering
\includegraphics[width=0.9\columnwidth]{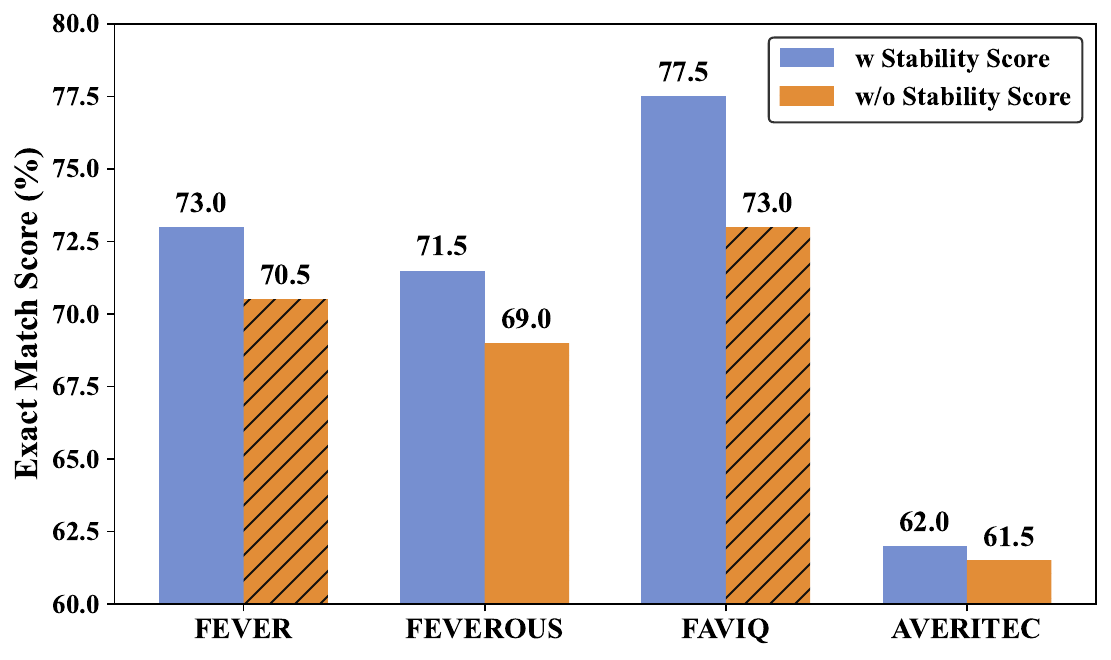}
\caption{Faithfulness and Answer Relevance scores across four datasets, based on the final predictions used for Exact Match evaluation, comparing models with and without scoring-based feedback.}
\label{fig5}
\end{figure}

\textbf{Number of Debate Rounds} We investigate how the number of debate rounds affects the overall performance of Tool-MAD. To analyze the effect of debate round, we conduct an experiment on the FEVER dataset. The results are shown in Figure ~\ref{fig3}. We observe that performance is lowest when only the Init Round (Round 1) is used. Accuracy gradually improves up to round 3 but slightly declines at round 4. Based on this observation, we set the threshold for the number of debate rounds to 3 in order to balance performance and efficiency.

This pattern suggests that although additional rounds help refine arguments and retrieve more relevant evidence, excessive debate depth may introduce unnecessary speculation or amplify intermediate reasoning errors. Such behavior aligns with observations in prior multi-step reasoning research, where prolonged interactions can lead to unstable or redundant argumentation. Therefore, three rounds provide sufficient opportunity for evidence refinement without causing degradation in the quality of the debate.

\vspace{0.2cm}

\textbf{Effect of Query Formulation} We conducted ablation experiments to evaluate the effectiveness of dynamic query formulation. In Tool-MAD, each agent can independently revise its query at every debate round to retrieve more relevant documents. To isolate the impact of this mechanism, we compare the standard Tool-MAD setup with dynamic query updates at each round (w/ Query Formulation) to a variant that uses only the initial claim for retrieval throughout all rounds (w/o Query Formulation). 
 
As shown in Figure~\ref{fig4}, dynamic query formulation improves performance on all datasets except F\textsc{a}VIQ, where the score remains unchanged. The most notable improvement is observed on FEVER (+2.0) and FEVEROUS (+2.5), followed by a modest gain on \textsc{AV}\textsc{eri}\textsc{TeC} (+1.0). These results demonstrate that formulating queries in response to the opponent’s argument helps uncover more relevant evidence and contributes to improved fact verification accuracy.

One possible explanation for this behavior is the difference in dataset characteristics. FEVER and FEVEROUS primarily contain entity-centric factual claims where emphasizing specific entities or relations during query reformulation leads to more targeted retrieval. In contrast, F\textsc{a}VIQ includes broader information-seeking questions that often require diverse or multi-hop evidence, meaning that the initial query already captures most of the relevant search space. As a result, further refinement has limited impact on F\textsc{a}VIQ, whereas datasets with more structured claim formulations benefit more from adaptive querying.

\vspace{0.2cm}

\begin{figure}[t]
\centering
\includegraphics[width=0.99\columnwidth]{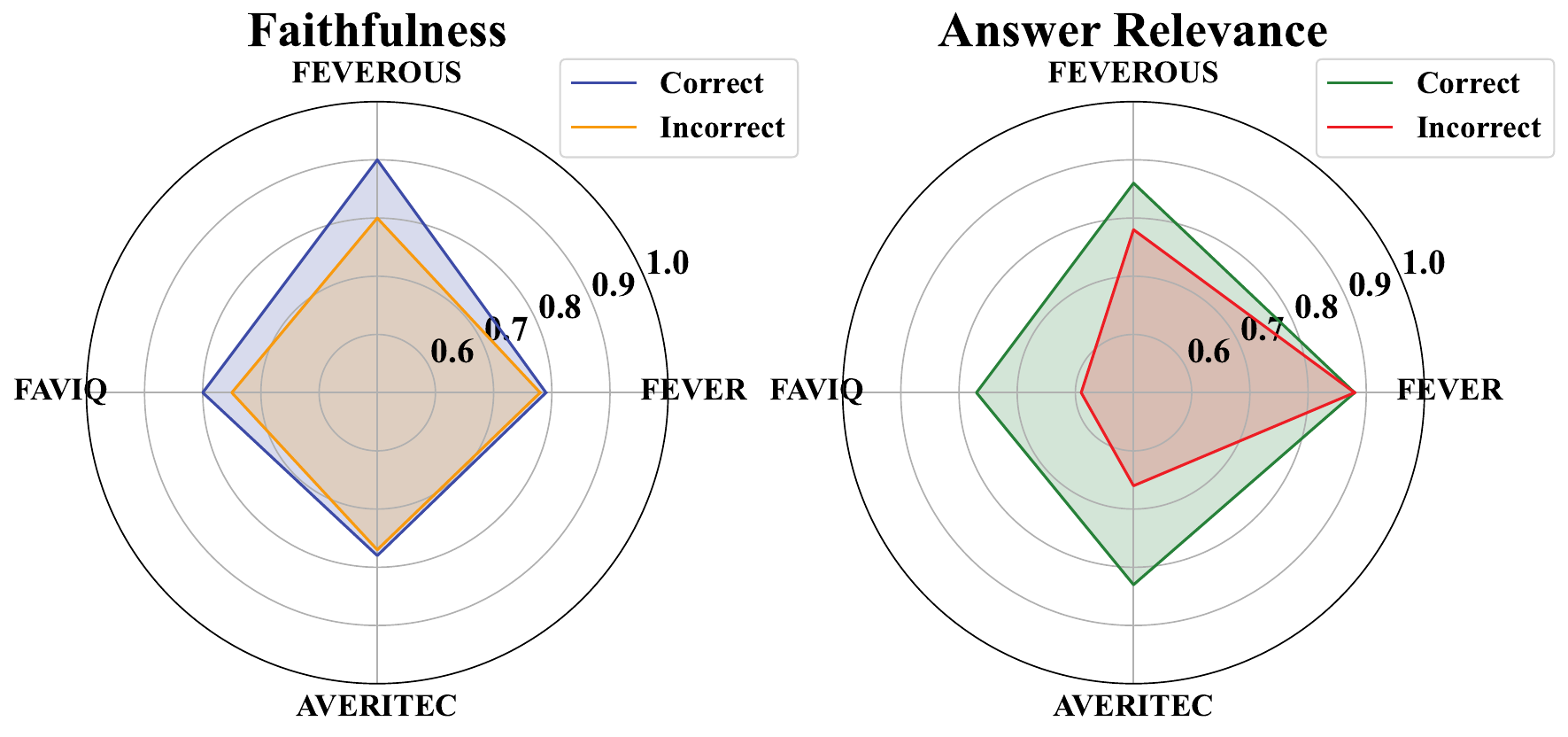} 
\caption{The figure presents the average Faithfulness and Answer Relevance scores for both correct and incorrect answers
across four datasets.}
\label{fig6}
\end{figure}

\textbf{Effect of Stability Score} We conducted ablation experiments to assess the effectiveness of incorporating faithfulness and answer relevance (stability score) as dynamic evaluation signals. In Tool-MAD, each agent’s response is evaluated at every debate round using these two metrics, and if either score falls below predefined thresholds, the round is considered inconclusive and is repeated. This mechanism enables the framework to dynamically filter out low-quality responses throughout the debate.

To isolate the impact of this score-based filtering, we compare the standard Tool-MAD configuration that utilizes faithfulness and answer relevance scores to guide debate progression (w/ Scoring Feedback) with a variant that skips score-based validation and progresses through all rounds regardless of response quality (w/o Scoring Feedback).

The results in Figure~\ref{fig5} highlight that scoring feedback consistently improves EM performance across all datasets, with the most significant gain observed on F\textsc{a}VIQ ($+4.5\%$). These findings suggest that faithfulness and answer relevance function not merely as evaluation metrics but as crucial control signals that help maintain factual consistency and question relevance throughout multi-round debates.

A notable aspect of this trend is that F\textsc{a}VIQ benefits particularly strongly from stability-based filtering. Because many F\textsc{a}VIQ questions are open-ended and sensitive to topical alignment, answers that deviate slightly from the intended question are often incorrect even when they contain factually valid information. In such cases, answer relevance becomes a highly predictive indicator of correctness, while faithfulness prevents unsupported reasoning from propagating across rounds. Together, these signals help stabilize the debate process and reduce both topic drift and hallucinated content.

\subsection{Analysis of Stability Score}
Figure~\ref{fig6} compares the average Faithfulness and Answer Relevance scores across four datasets, based on whether the agent’s prediction was correct or incorrect. In all datasets, responses classified as correct consistently exhibit higher scores on both metrics, indicating a strong correlation between these scores and the Exact Match (EM) accuracy. These findings suggest that the scores can be used not only as post-hoc evaluation metrics but also as internal signals for determining answer correctness or filtering responses in real-time systems.

In particular, the \textsc{FaVIQ} and \textsc{AV}\textsc{eri}\textsc{TeC} datasets show a pronounced gap in Answer Relevance scores between correct and incorrect responses, suggesting that alignment with the original user query plays a crucial role in predicting correctness for complex question answering tasks. These results highlight that Faithfulness and Answer Relevance serve as strong and reliable indicators for evaluating response quality and factual consistency. They can also contribute to quality control and final decision-making mechanisms in multi-agent debate systems.

\section{Discussion}

\subsection{Structural Effects of Query Rewriting}

One of the most salient patterns observed in our analysis is that the gradual query rewriting performed throughout the debate plays a crucial role in improving verification accuracy. Initial queries typically take a broad, holistic form that mirrors the surface structure of the claim, but as the debate progresses, the agents refine their queries by decomposing attributes, restructuring entity relations, or explicitly specifying missing details. For instance, in the Trinidad and Tobago Guardian ownership case, the first-round query asked about the newspaper’s overall “ownership history,” which retrieved only partial information. In later rounds, the agents rewrote the query to isolate the key relation: “the relationship between Guardian Media Limited and ANSA McAL,” which surfaced documents describing the hierarchical ownership chain. This shift transformed an initially ambiguous or contradictory prediction into a stable SUPPORTS decision. A similar phenomenon occurred in the Mel Ott RBI case. Early queries retrieved only season-level statistics, leading to NOT ENOUGH INFO (NEI), but rewriting the query into “total career RBIs” enabled retrieval of the exact aggregate value (1860), allowing the system to recover from partial evidence and converge on the correct label. These examples demonstrate that the multi-round query rewriting process is not merely repeated retrieval, but a structured mechanism that incrementally decomposes and reconstructs the claim, guiding the system toward increasingly precise evidence.

\subsection{Structural Effects of Stability scores}

A second notable characteristic identified in the analysis is the stabilizing effect of round-wise Stability scores, which evaluate each answer’s faithfulness (document–answer consistency) and answer relevance (question–answer consistency). It is common for the RAG-based answer and the Search-based answer to propose different labels for the same claim. Rather than selecting based on majority or repetition, the system computes stability scores for both answers at every round and tends to choose the label whose supporting answer maintains consistently higher semantic alignment with the retrieved evidence and the original question. For example, in the Stomatochaeta classification case, the RAG answer repeatedly predicted SUPPORTS while the Search answer predicted REFUTES. Across multiple rounds, however, the Search answer exhibited higher faithfulness scores and more stable question–answer relevance, prompting the model to adopt REFUTES as the final label. The Obispo first ascent case further illustrates this effect. Although the first round produced NEI due to insufficient evidence, subsequent rounds produced SUPPORTS answers with steadily increasing stability scores, eventually overriding the initial uncertainty and yielding a confident final decision. Because stability score directly measures whether an answer is grounded in retrieved evidence, it remains robust even under noisy retrieval conditions, assigning greater weight to evidence-aligned answers that consistently satisfy semantic constraints. Thus, stability score provides a meta-evaluative layer that re-assesses each candidate answer based on evidence quality and coherence, substantially enhancing convergence reliability compared to single-agent RAG systems.

\subsection{Multi-Agent Debate as a Structural Exploration Mechanism}

Finally, the analysis shows that the debate process exhibits an emergent capacity to explore structural properties of evidence, temporal shifts, and conflicting information even without an explicit moderator. A representative example is the Fort Myers Police Department case, where the Search tool retrieved up-to-date information indicating that the police chief had recently passed away, implying REFUTES, while the ground-truth label, anchored to an earlier timestamp, remained SUPPORTS. In such scenarios, the agents did not merely default to the labeled answer but instead engaged in a comparative evaluation of evidence recency, reliability, and contextual relevance. In other cases, the agents grounded their arguments in different retrieval sources and iteratively critiqued one another’s evidence, adjusting their interpretations in later rounds. This behavior suggests that the system is not simply producing parallel answers but is performing a form of quasi-critical reasoning that assesses the coherence, conflict, and incompleteness of available evidence. Such patterns indicate that multi-agent debate has the potential to evolve beyond static fact verification toward a richer inference system capable of handling temporal drift, evidence conflicts, and multi-faceted claims, conditions that more closely resemble real-world reasoning tasks.

\section{Conclusion}
In this work, we introduced Tool-MAD, a multi-agent debate framework that enhances factual verification by assigning distinct external tools to each agent and enabling adaptive, context-aware retrieval throughout the debate. Across diverse fact-verification and medical QA benchmarks, Tool-MAD achieved up to 35\% performance gains over existing debate systems, with ablation studies confirming the importance of tool diversity, dynamic query formulation, and our Stability Score, which functions as an internal control signal rather than a retrospective metric. These results demonstrate that combining structured debate with adaptive tool use significantly improves factual grounding and robustness. Tool-MAD’s stable performance across domains and retrieval configurations indicates strong potential for extension to richer tool ecosystems and more advanced judge models. Overall, Tool-MAD provides a unified and evidence-sensitive framework for transparent and reliable multi-agent reasoning in high-stakes factual verification tasks.

\bibliographystyle{IEEEtran}
\bibliography{main}

\newpage
\vspace{11pt}

\vfill

\end{document}